%% file: main_arXiv.tex
\documentclass{article}
\usepackage{spconf_modif,amsmath,graphicx,hyperref}
\usepackage{algorithm}
\usepackage{algpseudocode}
\usepackage{xcolor}
\usepackage{placeins}

\input{entete}

\title{\textsc{Nifty}: a Non-Local Image Flow Matching for Texture Synthesis} 
\name{Pierrick Chatillon,
    Julien Rabin, and
    David Tschumperlé
    \thanks{This work was partly funded by the Normandy Region through the IArtist excellence label project and benefited from computational resources provided by CRIANN.}}
\address{Normandie Univ, UNICAEN, ENSICAEN, CNRS, GREYC, F-14050 Caen, France}

\begin{document}
%
\maketitle

\input{sections/0_abstract}
\input{sections/1_intro}

\input{sections/2_method}

\input{sections/3_exp}

\input{sections/4_conclusion}

%
\FloatBarrier
\bibliographystyle{IEEEbib}
\bibliography{main}
\FloatBarrier
\input{sections/5_appendix}

\end{document}

%% file: entete.tex
\usepackage{tikz} 
\usepackage{multirow}
\usepackage{comment}
\usepackage{amssymb}
\usepackage{stmaryrd} 
\usepackage{bbm}
\usepackage[table]{xcolor}

\usepackage{subcaption}
\usepackage[margin=1in]{geometry}
\usepackage{array}

\newcommand{\old}[1]{\textcolor{red}{}}

\definecolor{ochre}{rgb}{0.8, 0.47, 0.13}
\definecolor{oceanboatblue}{rgb}{0.0, 0.47, 0.75}

\usepackage{hyperref}
\hypersetup{
    colorlinks=true,
    linkcolor=oceanboatblue,
    citecolor=ochre,
    filecolor=magenta,      
    urlcolor=cyan,
    pdftitle={Overleaf Example},
    pdfpagemode=FullScreen,
}

\def\x{{\mathbf x}}

\def\N{{\cal N}}

\def\UU{{\mathbb U}}

\def\P{{\mathcal P}}

%% file: sections/0_abstract.tex
\begin{abstract}

This paper addresses the problem of exemplar-based texture synthesis. We introduce \textsc{nifty}, a hybrid framework that combines recent insights on diffusion models trained with convolutional neural networks, and classical patch-based texture optimization techniques. \textsc{nifty} is a non-parametric flow-matching model built on non-local patch matching, which avoids the need for neural network training while alleviating common shortcomings of patch-based methods, such as poor initialization or visual artifacts. Experimental results demonstrate the effectiveness of the proposed approach compared to representative methods from the literature. Code is available at \url{https://github.com/PierrickCh/Nifty.git}.
\end{abstract}

%% file: sections/1_intro.tex
\section{Introduction and Related Work}
\label{sec:intro}

Image generative modeling has been a very active domain over the past decade, driven by a combination of theoretical and technical advances. 
This progress has led to the development of diverse generative modeling frameworks, many of which rely on the training of deep neural networks.

\paragraph*{Diffusion models}
Generative diffusion models (DMs)
\cite{diffusion,DDPM,SDE,DDIM,LatentDiffusion,RectifiedFlow,FlowMatching} 
have recently attracted significant attention for their ability to capture complex data distributions and generate high-quality samples, while benefiting from the stable training provided by conditional U-Net architectures.
Large pre-trained DMs, often conditioned on text prompts \cite{LatentDiffusion}, have also become widely used for inverse problems and image editing (\emph{e.g.}~\cite{song2023pseudoinverse}), and can be efficiently fine-tuned to address task-specific applications \cite{ControlNet}.

A major limitation of DMs lies in their sequential sampling process, formulated as a stochastic differential equation (SDE) \cite{diffusion,DDPM,SDE}, which requires a large number of small time steps during inference. 
This drawback has been alleviated by variants based on ordinary differential equations (ODE) \cite{karras2022elucidating}, where stochasticity is restricted to the initialization, including implicit models such as DDIM \cite{DDIM}, Rectified Flow (RF) \cite{RectifiedFlow}, and Flow Matching (FM) \cite{FlowMatching}. 
Inference can 
be significantly accelerated by distillation-based approaches, in which a student network is trained to reduce the number of required steps \cite{meng2023distillation,RectifiedFlow}.

\paragraph*{Texture Modeling with local patches}
Here we focus on exemplar-based texture synthesis, a task for which DMs are also gaining popularity, either through large pre-trained models \cite{zhou2024generating} or models specifically trained on a single image \cite{SIMuLDiTex_BMVC}. 
Before the emergence of such approaches, stationary texture modeling had been studied through a variety of statistical techniques designed to learn local representation. 
A particularly influential framework relevant to our method is the seminal work of Kwatra \emph{et al.} \cite{kwatra2005texture}, which introduced patch-based texture optimization (TO) which has since inspired single-image generative models based on patch representations, \emph{e.g.}~\cite{SinGAN,patex}.

In \cite{kwatra2005texture}, an image 
$x$ is synthesized by minimizing an energy function defined w.r.t to a reference image 
$u$, where each patch of the synthesized image must closely match a patch from $u$.
In practice, this non-convex energy minimization alternates between assigning each patch to its Nearest-Neighbor (NN) and averaging the overlapping patches at each pixel location. 
The method is, however, highly sensitive to random initialization and to hyper-parameters such as patch size and stride. 
When combined with a multi-scale strategy (coarse-to-fine synthesis), it can produce realistic samples, but often by replicating large regions of the reference image $u$, a behavior reminiscent of earlier methods that explicitly modeled such copy-paste effects \cite{graphcut}. 
These replicated regions are not always well aligned, frequently leading to discontinuities or blur, and fail to capture long-range correlations present in the reference. 
More recently, GAN-based approaches for single-image generation, such as \cite{SinGAN}, have been shown to exhibit the same limitations and can in fact be advantageously replaced by patch-based NN methods \cite{dropthegan}, without training a latent generative model.

\paragraph*{Creativity of generative models}
Since then, the question of creativity in generative models trained on large datasets has been closely examined. 
Studies such as \cite{webster2025multi} have shown that these models are prone to memorization, and that some training data can be extracted.
Recently, \cite{Analytic_creativity} demonstrated that an optimal diffusion models, \emph{i.e.} maximizing the likelihood of a training set boils down to memorizing it.
In practice, training a Convolutional Neural Networks (CNNs) to approximate the score driving the likelihood introduces a strong inductive bias: the training dataset is effectively processed as a collection of patches. %
This analysis allows for an explicit formula to model the learned diffusion by a CNN, where the score of a noisy sample at each time step essentially reduces to a mixture of Gaussians around each training patches.
While this approach is not practical for large-scale applications, it further highlights the connection between patch-based NN matching and diffusion processes.

\paragraph*{Contributions and Outline}
In this work, we propose \textsc{nifty}, an approximation of the explicit flow on patches for texture synthesis, that eliminates the need for neural network training and frame the TO algorithm~\cite{kwatra2005texture} 
as a temporal integration of the flow (Sec.~\ref{sec:method}).
In addition to coarse-to-fine synthesis with subsampling and aggregation strategies, we further introduce techniques, including top-$k$ sampling and memorization, to reduce the number of required steps while maintaining quality.
This approach offers several advantages shown in experiments (Sec.~\ref{sec:exp}): robustness to initialization, image quality, and speed.

%% file: sections/2_method.tex
\begin{figure*}[htb]
    \centering
    \setlength{\tabcolsep}{1pt}  
    \renewcommand{\arraystretch}{1}  
    \begin{tabular}{@{} c c c @{}}
        \includegraphics[width=0.11\linewidth]{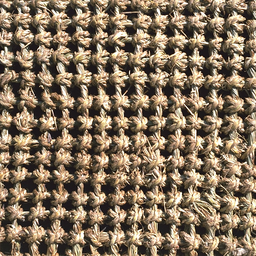} &
        \includegraphics[width=0.44\linewidth]{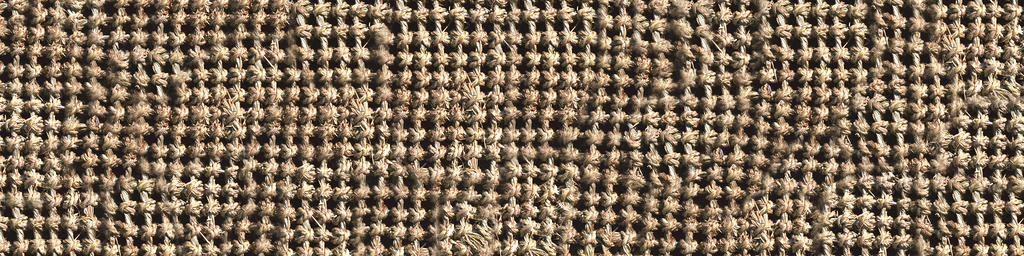} &
        \includegraphics[width=0.44\linewidth]{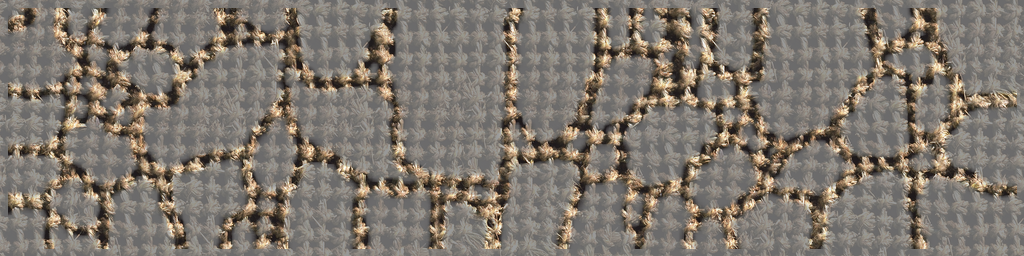}\\
        \includegraphics[width=.11\linewidth]{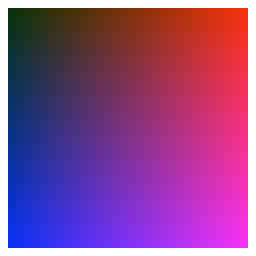} &
        \includegraphics[width=0.44\linewidth]{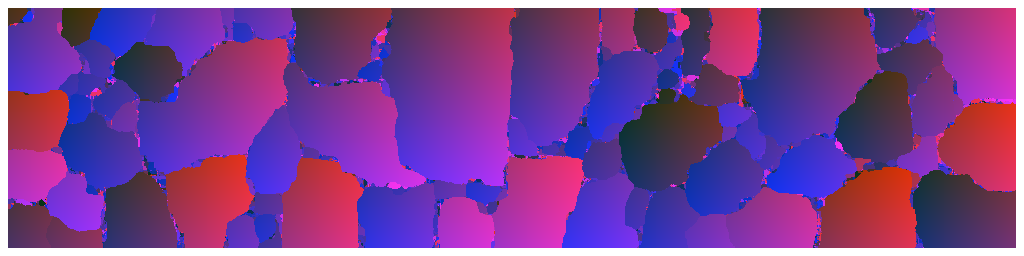} &
        \includegraphics[width=0.44\linewidth]{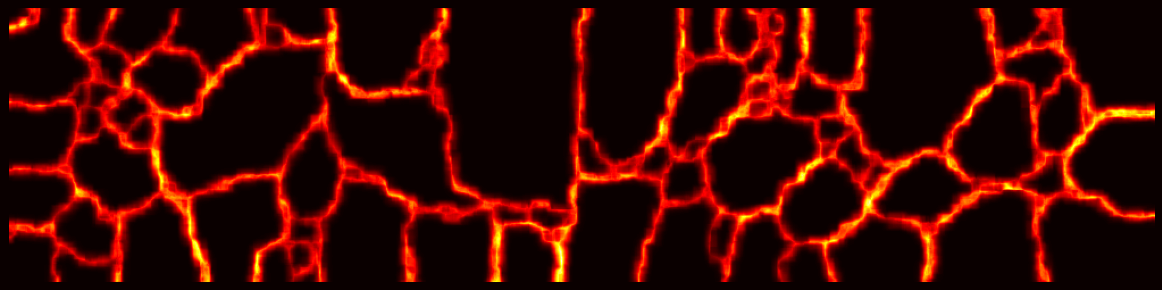} 
    \end{tabular}
    \vspace*{-3mm}
    \caption{\small \textbf{Creativity in synthesis.} 
    Distance to the Nearest Neighbor (NN) patch in the training set is used to highlight novel structures.
    From left to right, first row: example image, generated sample, 
    new structures;
    second row: pixel colormap, NN colormap, NN distance.
    This illustrates how the generative model manages to stitches local copies of the original texture by creating unseen yet likely structures.
    }
    \label{fig:novelty}
\end{figure*}
\section{Method}
\label{sec:method}

The Flow Matching (FM) framework, as introduced in~\cite{FlowMatching,RectifiedFlow} consists in sampling a complex data distribution by integrating a velocity field, starting from Gaussian samples. 
This velocity field is often approximated by a CNN. 
As discussed in the introduction, the fact that CNNs are used to compute the flow (or diffusion denoising steps) introduces the inductive bias of equivariance. As a consequence, computing the velocity field at a noisy image can be understood as applying the same local estimation to all patches.
Subsequently, we will detail how to compute explicitly the flow on patch distributions (Sec.~\ref{subsec:FM_patch}) and then approximate it efficiently (Sec.~\ref{subsec:Nifty_alg}) without neural networks for texture synthesis.

\subsection{Non-Local Image Flow-Matching}
\label{subsec:FM_patch}

\paragraph*{Notations and formal setting} We consider the collection of training patches $\P := \{\phi\}$ from the reference image $u$ with given patch size and stride.
Note that to avoid border effect in practice,
patches across boundaries are simply discarded. 

We denote by $\N(\mu,\sigma^2)$ the isotropic multivariate Gaussian probability law with mean $\mu$ and standard deviation $\sigma$, and its density by $g_{\mu,\sigma}(x) = g_{0,1}(\tfrac{x-\mu}{\sigma})
\propto \exp{-\tfrac{(x-\mu)^2}{2\sigma^2}}$.

\paragraph*{FM formulation} 
Starting from a random (patch) sample $\psi_0 \sim \N (0,I)$ at time step $t=0$, the FM framework consists during inference in solving the following ODE for $t \in (0,1]$:
{\setlength{\abovedisplayskip}{5pt}
 \setlength{\belowdisplayskip}{5pt}%
\begin{equation}
\label{eq:ode}
    \partial_t \psi_t = v(\psi_t,t)
\end{equation}}%
where $v$ is the velocity of the flow
such that generated samples $\psi_1 \sim \UU (\P)$ match the training distribution.

In \cite{FlowMatching}, authors argue that while the problem of defining such a vector field is intractable, different vector fields are solution.
They show however that the problem can be tackled by considering an affine parametrization of the conditional flow $\psi_t$  given a target sample $\phi$, using 
{\setlength{\abovedisplayskip}{5pt}
 \setlength{\belowdisplayskip}{5pt}%
\begin{equation}\label{eq:affine_map}
    \psi_t = t \phi + (1-t) \psi_0.
    \vspace*{-1em}
\end{equation}}

\paragraph*{Training}
In practice, the corresponding vector field is learned using a neural network $v_\theta$ trained to minimize 
%
{\setlength{\abovedisplayskip}{5pt}
 \setlength{\belowdisplayskip}{5pt}%
\begin{equation}\label{eq:FM_training}
    \min_{\theta} 
    \mathbb{E}_{\substack{
        \phi \sim \mathbb{U}(\mathcal{P}),
        z \sim \mathcal{N}(0,I) \\
        t \sim \mathbb{U}([0,1])
    }} 
    \left\|
    v_\theta(\phi_t, t) - (\phi - z) \right\|^2
    \vspace*{-1em}
\end{equation}}

\paragraph*{Interpretation}
Optimizing \eqref{eq:FM_training} on all possible functions at each time step is a barycentric problem which solution writes
{\setlength{\abovedisplayskip}{5pt}
 \setlength{\belowdisplayskip}{5pt}%
\begin{equation}\label{eq:velocity_expectation}
v(\psi,t)=\mathbb{E}_{\substack{
        \phi \sim \mathbb{U}(\mathcal{P})\\
        z \sim \mathcal{N}(0,I)
    }} \Big ( 
\phi - z \Big | \psi = \phi t + z (1-t) \Big)
.
\end{equation}}%
Besides, the push-forward of the gaussian latent prior conditionally to a data point $\phi$ using such the affine mapping \eqref{eq:affine_map} is a conditional Gaussian probability distribution
{\setlength{\abovedisplayskip}{5pt}
 \setlength{\belowdisplayskip}{5pt}%
\begin{equation}\label{eq:conditional_gaussian}
p(\psi_t|\phi) = g_{t\phi, (1-t)^2} (\psi_t).
\vspace*{-1em}
\end{equation}}

\paragraph*{Closed-form solution}
Combining expressions \eqref{eq:velocity_expectation} and \eqref{eq:conditional_gaussian} shows that one needs to sample a latent variable $z = \frac{\psi - t\phi}{1-t}$ to satisfy the condition $\psi = \phi t + z (1-t)$.
Using the Bayes' rule, 
the computation of the conditional expectation \eqref{eq:velocity_expectation} then reduces to a Gaussian mixture centered around each training data point
{\setlength{\abovedisplayskip}{5pt}
 \setlength{\belowdisplayskip}{5pt}%
\begin{equation}\label{eq:NL-patch_velocity}
    \begin{split}
        v(\psi,t)
            &= \mathbb{E}_{\substack{\phi \sim \mathbb{U}(\P)}} 
            \Big ( \tfrac{\phi - \psi}{1-t} \Big | \psi \Big)
        \\
            &= \frac{1}{1-t} \int_\phi (\phi - \psi)
            \frac{p(\psi | \phi)}{p(\psi)} dp(\phi) 
        \\
            &= \frac{1}{1-t} \frac{1}{|\P|p(\psi)} \sum_{\phi \in \P} 
            (\phi - \psi) p(\psi | \phi)
        \\
            &= \frac{1}{1-t}  \sum_{\phi \in \P} \big(\phi-\psi \big) \omega_{\psi, t}(\phi) 
            \\
            & \text{with  } \omega_{\psi, t}(\phi)=
     \frac{g_{t\phi, (1-t)^2} (\psi)}{\sum_{\phi'} g_{t\phi', (1-t)^2} (\psi)}
    \end{split}
\end{equation}}

%
Note that this formula is analogous to \cite[Eq. 5]{Analytic_creativity} derived for an optimal equivariant score machine.
Interestingly, this formula is also reminiscent of the Non-Local Means \cite{NLM} where noisy patches are aggregated for image denoising applications.

When starting the flow at $t=0$, it boils down to the direction towards the empirical mean: $\bar{\phi} - \psi$.
Unfortunately, this formula becomes prohibitively expensive to compute for large numbers of patches, since all the dataset has to be used to compute the velocity at each step, for each patch under synthesis.

\subsection{\textsc{nifty} Algorithm}
\label{subsec:Nifty_alg}

The \textsc{nifty} algorithm, 
stands for Non-local Image Flow-matching Texture sYnthesis.
It operates the previously introduced FM on patches
in a \textit{multi-scale} manner. 
At each resolution level, a noisy version of the upsampled signal from the previous level is denoised by solving the FM ODE \eqref{eq:ode} based on the proposed approximation \eqref{eq:NL-patch_velocity}.
The computational cost is reduced through two main approximations.

\paragraph*{Top-$k$ NN}
Instead of evaluating the full weighted sum in equation~\eqref{eq:NL-patch_velocity}, we restrict computation to the $k$ patches $\{\hat{\phi}_i\}_{i \leq k}$ with the largest weights, which are the $k$-NN of $\tfrac{1}{t}\psi$. 
The corresponding weight approximation is given by:
{\setlength{\abovedisplayskip}{5pt}
 \setlength{\belowdisplayskip}{5pt}%
\begin{equation}
    \label{eq:omega_k}
    \hat{\omega}_{\psi, t}(\hat{\phi}_j)=\frac{g_{t\hat{\phi}_j, (1-t)^2}(\psi)}{\sum_{i=1}^k g_{t\hat{\phi}_i, (1-t)^2}(\psi)}
    \vspace*{-1em}
\end{equation}}

\paragraph*{Memory}
To further reduce computation, we consider only a subset $\{\phi^r\} \subset \mathcal{P}$ of the training patch set, where $|\{\phi^r\}| = r|\mathcal{P}|$ for some sampling ratio $r \in (0,1]$. 
To mitigate the degradation of synthesis caused by this subsampling, we introduce a \textit{memory function}: 
$m : \llbracket1,k\rrbracket \rightarrow \llbracket1,|\mathcal{P}| \rrbracket$
, which retains the indices of the 
$k$-NN of each synthesized patch $\psi$ across iterations $t$, only updated with better matches.

\paragraph*{Pseudo-code}
Combining the two approximations, the flow for a patch $\psi$ at time $t$ and each scale is computed as follows:
\begin{enumerate}
    \item \textbf{Sampling}: Select a subset $\{\phi^r\}$ of cardinality $r|\mathcal{P}|$.
    \item \textbf{Neighborhood search}: Find $\{\hat{\phi}\}$, the $k$-NN of $\psi$ within the set $\{\phi^r\} \cup \{\phi_{m(i)} \,|\, 1 \leq i \leq k\}$.
    \item \textbf{Memory update}: Store new indices of the $k$-NN in $m$.
    \item \textbf{Weight computation}: Use \eqref{eq:omega_k} to compute $\hat{\omega}_{\psi, t}(\hat{\phi}_j)$.
    \item \textbf{Velocity estimation}: Compute the weighted velocity:
    {\setlength{\abovedisplayskip}{5pt}
 \setlength{\belowdisplayskip}{5pt}%
    \begin{equation}\label{eq:topk_FM}
        \hat{v}(\psi, t) = \frac{1}{1 - t} \sum_{i=1}^k \left(\hat{\phi}_i - \psi \right) \, \hat{\omega}_{\psi, t}(\hat{\phi}_i)
    \end{equation}}
    \item \textbf{Flow update:} an Euler scheme is used to update the patch $\psi$ using \eqref{eq:topk_FM} with ODE \eqref{eq:ode};
    \item \textbf{Stride:} to accelerate synthesis, 
    we sample patches from the current synthesis with a spatial stride; 
    \item \textbf{Aggregation:} overlapping patches at a pixel are averaged using a spatial gaussian kernel.
\end{enumerate}

\paragraph*{Link between \textsc{nifty} and TO}

As recalled in Sec.~\ref{sec:intro}, TO consist in alternating between patch-matching and aggregation, at different image resolutions and for different patch sizes (whereas we fix the patch size in \textsc{nifty}).
As such, TO resembles to the proposed approach when selecting a single NN patch ($k=1$) and performing a single time step ($T=1$).
Direct replacement with the nearest neighbor is a violent operation, which makes this algorithm sensible to initialization.
By merging the flows, rather than the nearest matches, our approach has a more stable behavior and does not necessitate to go through different patch sizes at each resolution.
As $t$ tends to 1, the weights $\omega_{\psi, t}$ (which sum to 1) tends to the indicator of the nearest neighbor, and thus the flow points directly from $\psi$ to its nearest neighbor. In that way, the final steps of our algorithm resemble the steps of TO.

It is worth noting that this non-local interpretation of FM for the TO algorithm is related to the work of \cite{duval_gretsi_kwatra_diffusion}, which also derives an explicit formula for DDPM. 
A key difference, however, is that their formulation restricts the score approximation to local patch NN matching, \emph{i.e.}, comparing only patches at identical spatial locations across registered images.

%% file: sections/3_exp.tex
\section{Experiments}
\label{sec:exp}

We first present experimental results (comparison, ablation) of the proposed \textsc{nifty} approach using patch based representation.
We consider in all experiments a fixed patch-size of $16$ px with a sampling stride of $4$ px, at 4 different scales.
Recall that TO algorithm makes use of several patch size ($32,16,8$ px).
Then, we show how the proposed framework can be easily transposed to latent presentations, \emph{e.g.} latent DM~\cite{LatentDiffusion}. 

\begin{figure}[tb]
    \centering
    \includegraphics[width=\linewidth]{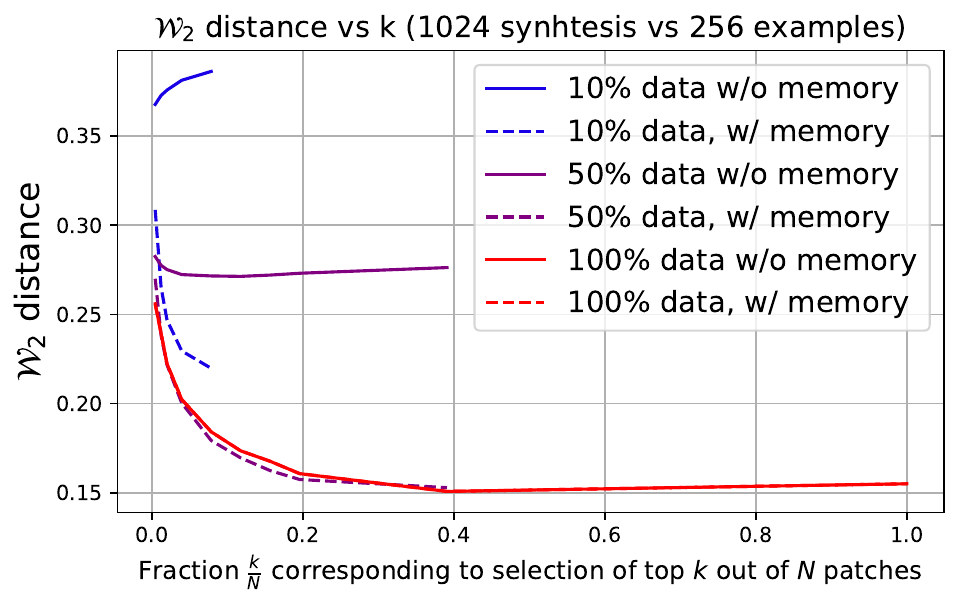}
    \vspace*{-6mm}
    \caption{\small \textbf{Effect of $k$-NN memory.} See text for details.}
    \label{fig:k_rs}
\end{figure}

\paragraph*{Pixel-space \textsc{nifty}}
Figure~\ref{fig:k_rs} illustrates the effect of introducing memory in combination with top-$k$ patch selection for improving the flow approximation. 
The flow quality is evaluated using the Wasserstein distance between patch distributions, computed on non-overlapping patches to avoid the influence of aggregation regularization. As expected, increasing $k$ yields a better fidelity. Adding memory allows for comparable synthesis with a fraction of the computational cost.

\begin{figure}[tb]
    \centering
    \includegraphics[width=0.48\linewidth,trim=0 1.5cm 0 0, clip]{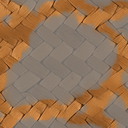}
    \includegraphics[width=0.48\linewidth,trim=0 1.5cm 0 0 , clip]{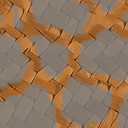}
    \vspace*{-2mm}
    \caption{\small \textbf{Artifacts.} Left: synthesis by \textsc{nifty}. Right: synthesis by TO. The synthesized patches not equal to their nearest neighbor from the reference are highlighted, as detailed in Fig.\ref{fig:novelty}.}
    \label{fig:artifacts}
\end{figure}

Figure~\ref{fig:novelty} shows an example of texture synthesis with the \textsc{nifty} algorithm.
It illustrates how the proposed techniques drive the proposed FM to copy local regions of the training image (as demonstrated in \cite{Analytic_creativity}), while creating new unseen yet likely structures.
A comparison in Fig.~\ref{fig:artifacts} to the TO algorithm underlines the interest of the proposed approach which avoid strong artifacts between copied regions.


To confirm this visual inspection, a quantitative comparison with TO is proposed in Table~\ref{tab:metrics}, computed from 10 synthesis (at 512 px res.) for 12 reference images (at 256 px). For reference, a small U-Net ($.7M$ parameters) is trained for each reference image to solve \eqref{eq:FM_training}, highlighting the interest of the proposed approach in terms of quality and speed.
\begin{table}[htb]
\centering
\resizebox{1.\linewidth}{!}{
\begin{tabular}{|l||c|c|c|c|}
\hline
\textbf{Method} & Gram $\downarrow$ \cite{gatys2015texture} & SIFID $\downarrow$ \cite{SinGAN}& Autocorrelation $\downarrow$ & Time (s) \\
\hline
\textsc{nifty} ($T=15$, $k=5$)& 3 601 &0.28&85.9  & 0.70 \\
TO  &          12 676 &0.76&431.8 & 1.92 \\
U-Net & 17034&0.54&133.1  & 600 (train)\\
($T=15$) & & & & 0.13 (eval.)
\\
\hline
\end{tabular}}
\vspace*{-3mm}
\caption{\small \textbf{Quantitative analysis.}  Comparison using 3 metrics.}
\label{tab:metrics}
\end{table}

\begin{figure}[tb]
    \centering
    \includegraphics[width=0.49\linewidth]{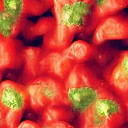}
    \includegraphics[width=0.49\linewidth]{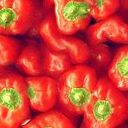}
    \vspace*{-2mm}
    \caption{\small \textbf{Mono-scale \textsc{nifty} vs FM network synthesis}  
    Left: using our explicit approximation ($k=10$). 
    Right: using a U-net approximation of FM. 
    Both methods start from the same noise realization.}
    \label{fig:nn_vs_fm}
\end{figure}

Figure~\ref{fig:nn_vs_fm} provides a visual comparison between image synthesis with \textsc{nifty} and with a U-Net approximation of the flow.  Both synthesis are similar, and use the same random Gaussian initialization, demonstrating that they indeed approximate the same flow. To do so \textsc{nifty} is restricted to the finest scale for generation, degrading the results.

\begin{figure}[tb]
    \centering
    \includegraphics[width=.49\linewidth]{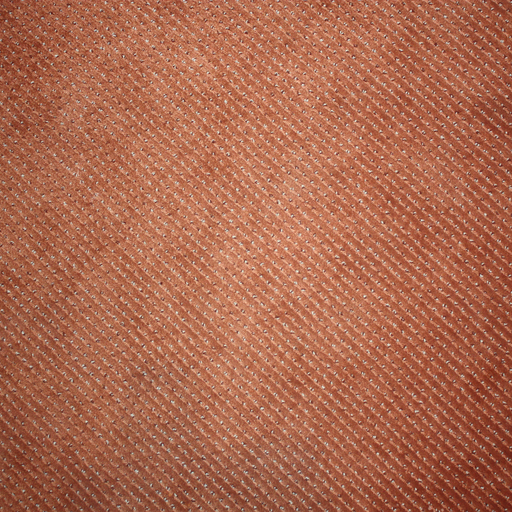}
    \includegraphics[width=.49\linewidth]{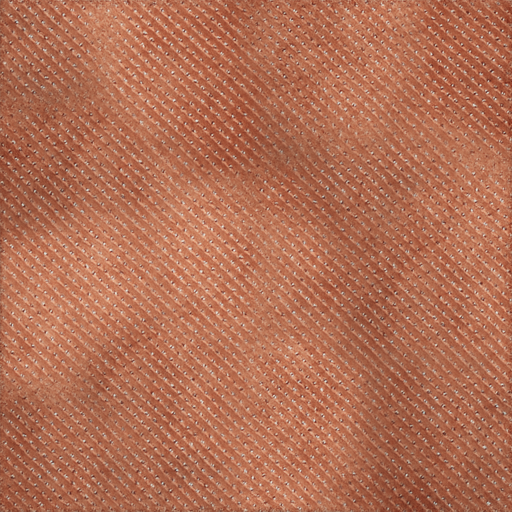}
    \vspace*{-2mm}
    \caption{\small \textbf{Latent \textsc{nifty} model} Left: reference image. Right: Our synthesis algorithm performed in the latent space of an auto-encoder}
    \label{fig:AE_flow}
\end{figure}

\paragraph*{Latent-Nifty} 
Figure~\ref{fig:AE_flow} presents a synthesis result obtained by combining \textsc{nifty} with an autoencoder pre-trained on the example patch distribution, highlighting the potential of the proposed method to be applied in latent spaces.

%% file: sections/4_conclusion.tex
\section{Conclusion}
\label{sec:ccl}

We introduced a generative model based on Non-Local Flow-Matching inspired from recent insights on deep convolutional generative diffusion models.
Revisiting the texture optimization algorithm from \cite{kwatra2005texture}, we show that we can generate efficiently large random textures from a single example without the need to train a neural network, while avoiding the major pitfalls of the original approach, such as visual artefacts and sensitivity to initialization.

In addition to exploring patch flow matching in suitable latent spaces, future work includes the integration of more sophisticated inductive biases. For instance, attention modules could be modelled by non-local patch aggregation.

\vfill\pagebreak

%% file: sections/5_appendix.tex
\appendix

\section{Detailed algorithm}
A more in-depth view of the \textsc{nifty} algorithm is described in Table~\ref{alg:patchfm}.
\input{sections/2_algo}

\section{Interpolation experiments}
We display three approaches to blending texture exemplars using our patch-based method: distribution-level blending, pixel-level blending, and spatial interpolation of blending weights.

\begin{figure}[htbp]
    \centering
    \includegraphics[width=\linewidth]{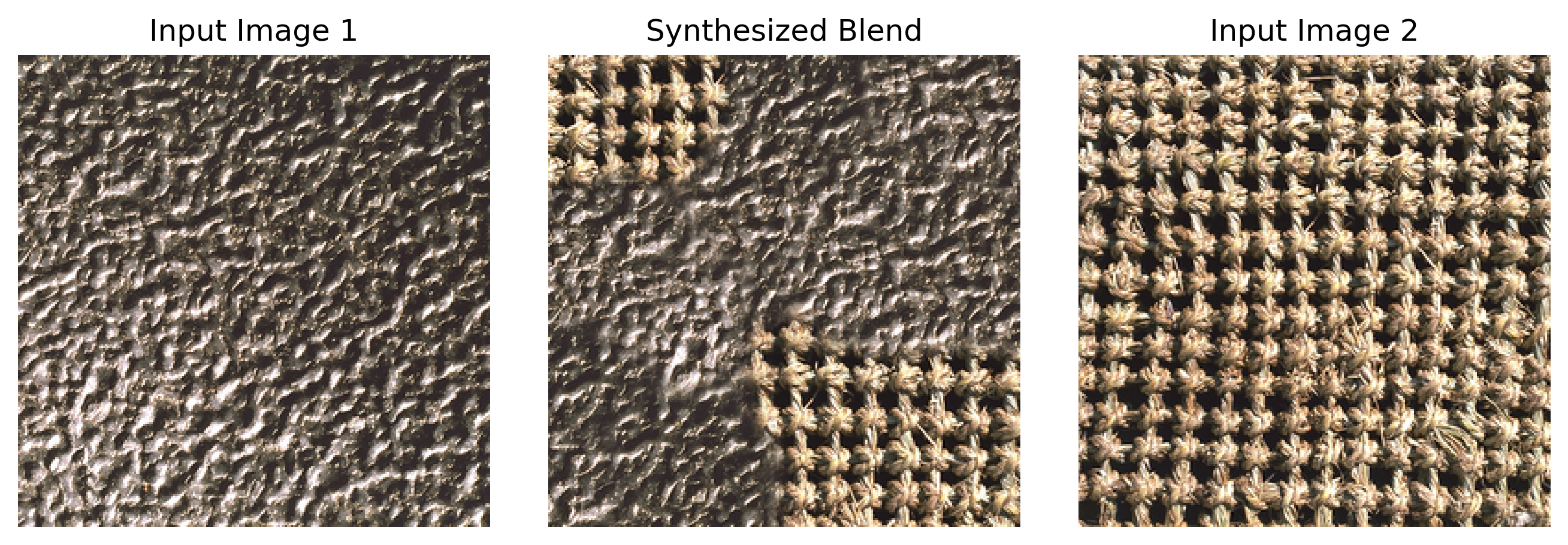}
    \caption{Distribution-level blending. Patch matching is performed over the union of both input texture patch sets, enforcing statistical homogeneity.}
    \label{fig:dist_blending}
\end{figure}

\textbf{Distribution-level blending} (Fig.~\ref{fig:dist_blending}) performs patch matching using the union of patches from both inputs. This yields an image with half of the locations looking like one exemplar.  

\begin{figure}[htbp]
    \centering
    \includegraphics[width=\linewidth]{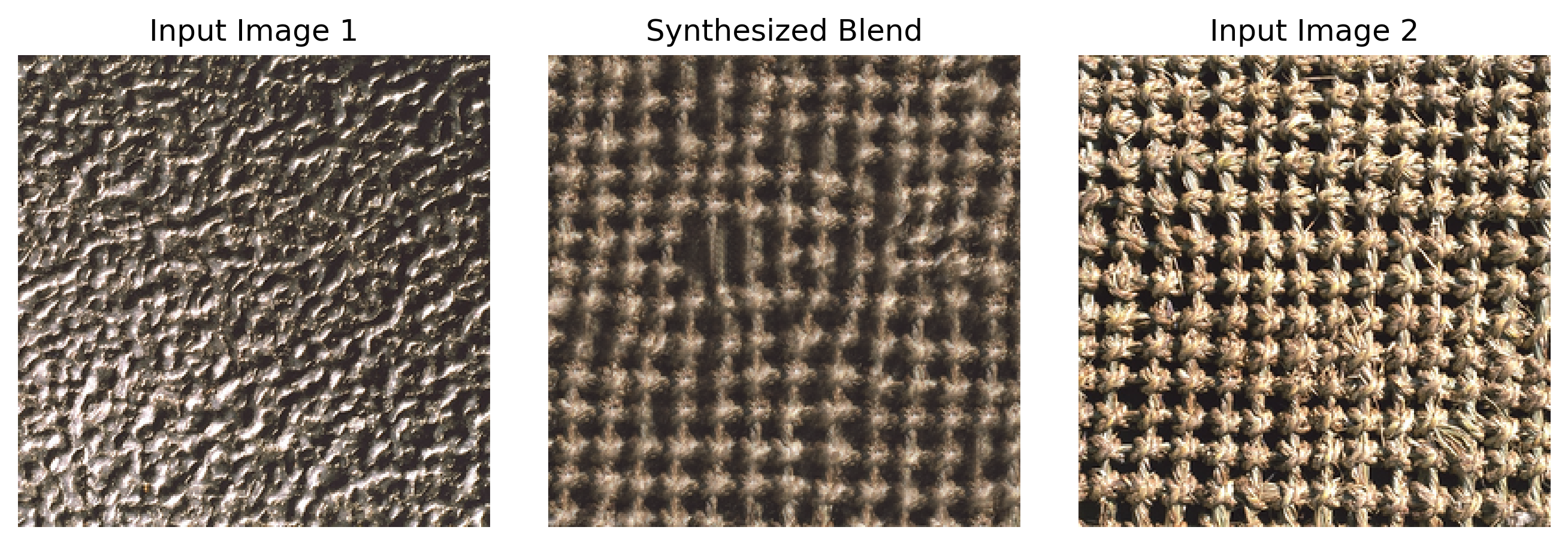}
    \caption{Pixel-level blending. The input textures (left, right) are synthesized by computing patch matches and interpolating their synthesis velocities with a global blending parameter $\alpha = 0.5$.}
    \label{fig:pixel_blending}
\end{figure}

\textbf{Pixel-level blending} (Fig.~\ref{fig:pixel_blending}) instead computes patch matches separately for each input and interpolates their synthesis velocities linearly with a global $\alpha$ weight.  

\begin{figure}[htbp]
    \centering
    \includegraphics[width=\linewidth]{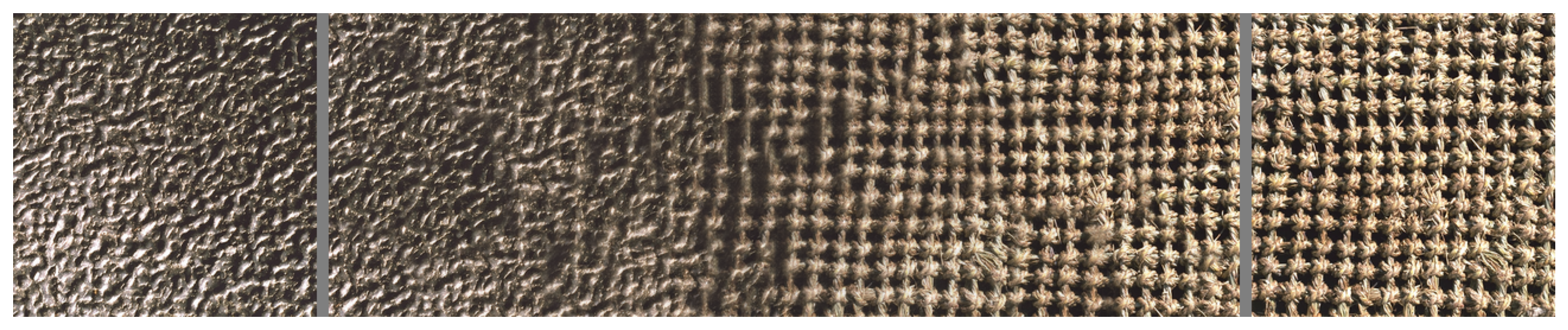}
    \caption{Spatial interpolation. A spatially-varying $\alpha$ map is used to transition smoothly between input textures, enabling region-aware synthesis.}
    \label{fig:spatial_blending}
\end{figure}

Finally, \textbf{spatial interpolation} (Fig.~\ref{fig:spatial_blending}) uses a spatially-varying $\alpha$ map, enabling smooth transitions across regions and better spatial control over blending.

\section{More results}
\begin{figure*}[htbp]
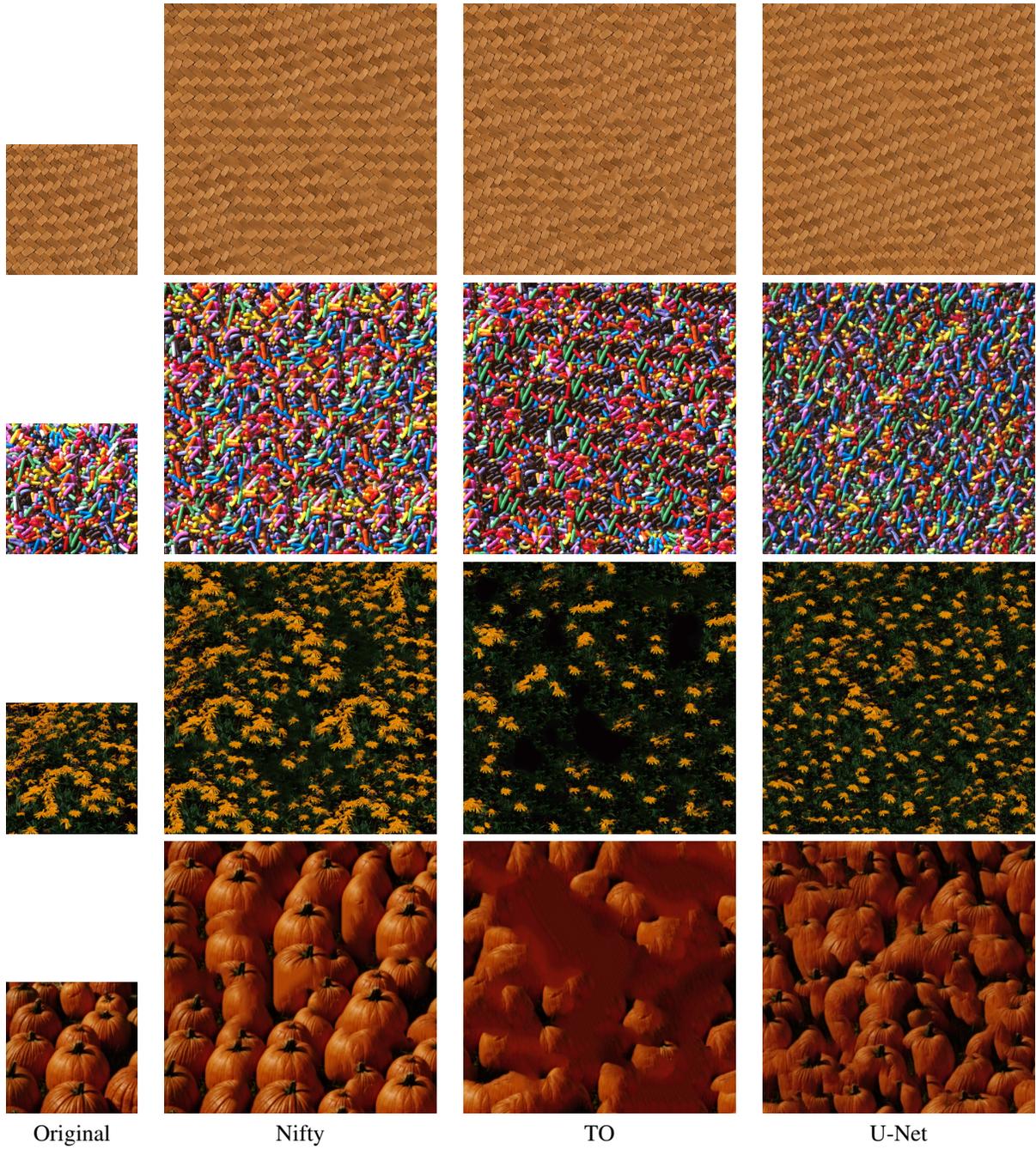

  \centering
  \newcommand{\mysize}{0.25\textwidth}
  \newcommand{\mysizemini}{0.12\textwidth}
  \resizebox{1.\linewidth}{!}{
  \begin{tabular}{*{4}{c}} 
    \def\rows{}%
    \foreach \x in {1,2,3,4}{%
      \xdef\rows{\rows
        \noexpand\includegraphics[width=\mysizemini]{figs/comp/base/\x.png} &
        \noexpand\includegraphics[width=\mysize]{figs/metrics/\x/nifty/0.png} &
        \noexpand\includegraphics[width=\mysize]{figs/metrics/\x/kwatra/0.png} &
        \noexpand\includegraphics[width=\mysize]{figs/metrics/\x/FM/0.png}
        \noexpand\\%
      }%
    }%
    \rows
    Original & Nifty & TO & U-Net
    \\
  \end{tabular}
  } 
  \caption{Comparison of synthesis across methods and images from the dataset used to compute average score in Table~\ref{tab:metrics}.}
  \label{fig:more1}
\end{figure*}

For all methods and for 4 images among the 12 of the base, we show in Fig.~\ref{fig:more1},\ref{fig:more2} and \ref{fig:more3} one of the 10 synthesis used to compute the values in Table \ref{tab:metrics}

\begin{figure*}[htbp]
  \centering
  \newcommand{\mysize}{0.25\textwidth}
  \newcommand{\mysizemini}{0.12\textwidth}
  \resizebox{1.\linewidth}{!}{
  \begin{tabular}{*{4}{c}} 
    \def\rows{}%
    \foreach \x in {5,6,7,8}{%
      \xdef\rows{\rows
        \noexpand\includegraphics[width=\mysizemini]{figs/comp/base/\x.png} &
        \noexpand\includegraphics[width=\mysize]{figs/metrics/\x/nifty/0.png} &
        \noexpand\includegraphics[width=\mysize]{figs/metrics/\x/kwatra/0.png} &
        \noexpand\includegraphics[width=\mysize]{figs/metrics/\x/FM/0.png}
        \noexpand\\%
      }%
    }%
    \rows
    Original & Nifty & TO & U-Net
    \\
  \end{tabular}
  } 
  \caption{Comparison of synthesis across methods and images from the dataset used to compute average score in Table~\ref{tab:metrics}.}
  \label{fig:more2}
\end{figure*}

\begin{figure*}[htbp]
  \centering
  \newcommand{\mysize}{0.25\textwidth}
  \newcommand{\mysizemini}{0.12\textwidth}
  \resizebox{1.\linewidth}{!}{
  \begin{tabular}{*{4}{c}} 
    \def\rows{}%
    \foreach \x in {9,10,11,12}{%
      \xdef\rows{\rows
        \noexpand\includegraphics[width=\mysizemini]{figs/comp/base/\x.png} &
        \noexpand\includegraphics[width=\mysize]{figs/metrics/\x/nifty/0.png} &
        \noexpand\includegraphics[width=\mysize]{figs/metrics/\x/kwatra/0.png} &
        \noexpand\includegraphics[width=\mysize]{figs/metrics/\x/FM/0.png}
        \noexpand\\%
      }%
    }%
    \rows
    Original & Nifty & TO & U-Net
    \\
  \end{tabular}
  } 
  \caption{Comparison of synthesis across methods and images from the dataset used to compute average score in Table~\ref{tab:metrics}.}
  \label{fig:more3}
\end{figure*}

%% file: sections/2_algo.tex
\begin{algorithm}
\small
\caption{\textsc{nifty} Patch Flow Matching}
\label{alg:patchfm}
\begin{algorithmic}[1]
    \Statex \textbf{Input:} Exemplar image $u$
    \Statex \textbf{Hyperparameters:} Number of scales $S$, patch size $p$, number of neighbors $k$, renoising factor $\gamma$, subsampling ratio $r$, Number of timesteps T.
    \Statex \textbf{Output:} Synthesized image 
    \State Initialize synthesis image $x\sim \N(0,1)$ with noise
    \State $\{t_i\} \gets $  $T$ linear timesteps between 0 and 1 
    \For{$s = S-1$ to $0$}
        \State Resize $u$ to resolution of scale $s$
        \State Extract training patches $\phi$
        \If{$s>S-1$}
            \State Upsample previous synthesis $x$
            \State \textit{Renoising:} $x \gets \gamma x + (1 - \gamma) \varepsilon, \varepsilon \sim \N(0,1)$ 
            \State $\{t_i\} \gets$  $T$ linear timesteps between $ \gamma$ and 1 
        \EndIf
        \For{$i = 1$ to $T$}
            \State $\{\psi\} \gets$ patches of $x$, using stride $\tfrac{p}{4}$
            \State Select $\{\phi^r\}$, uniformly sampling $\{\phi\}$
            \State Retrieve $\{\phi_{m(i)}, 1\leq i \leq k\}$
            \State Compute $\hat{\phi}_{i\leq k}$, the $k$-NN of $\frac{1}{t}\psi$ from $\{\phi^r\} \cup \{\phi_{m(i)}, 1\leq i \leq k\}$, update $m$
            \State Compute weights $\hat{\omega}_{\psi^r_i, t}(\hat{\phi})$
            \State $\hat{v}(\psi,t) \gets \frac{1}{1-t}  \sum_{i=1}^k \big(\hat{\phi}_i-\psi \big) \hat{\omega}_{\psi, t}(\hat{\phi}_i) $
            \State $v_{agg}(x,t) \gets$ Aggregation of $\hat{v}(\psi,t)$
            \State $\x \gets \x + (t_{i+1}-t_i) v_{agg}(x,t)$

        \EndFor
    \EndFor
    \State \Return $x$
\end{algorithmic}
\end{algorithm}